\documentclass{article}

\usepackage[english]{babel}

\usepackage[letterpaper,top=2cm,bottom=2cm,left=3cm,right=3cm,marginparwidth=1.75cm]{geometry}

\usepackage{amsmath}
\usepackage{lscape}
\usepackage{rotating}
\usepackage{multirow}
\usepackage{graphicx}
\usepackage{booktabs}
\usepackage{adjustbox}
\usepackage{array}
\usepackage[colorlinks=true, allcolors=blue]{hyperref}

\title{Was that Sarcasm?: A Literature Survey on Sarcasm Detection}
\author{Harleen Kaur Bagga, Jasmine Bernard, Sahil Shaheen, Sarthak Arora}

\author{Harleen Kaur Bagga* \\\href{mailto:har.bagga8@gmail.com}{har.bagga8@gmail.com}
\and Jasmine Bernard* \\\href{mailto:jasmine0696@gmail.com}{jasmine0696@gmail.com}
\and Sahil Shaheen* \\\href{mailto:sahilshaheen14@gmail.com}{sahilshaheen14@gmail.com}
\and Sarthak Arora* \\\href{mailto:sarthakvaror@gmail.com}{sarthakvarora@gmail.com}
}

\begin{document}
\maketitle

\begin{abstract}
Sarcasm is hard to interpret as human beings. Being able to interpret sarcasm is often termed as a sign of intelligence, given the complex nature of sarcasm. Hence, this is a field of Natural Language Processing which is still complex for computers to decipher. This Literature Survey delves into different aspects of sarcasm detection, to create an understanding of the underlying problems faced during detection, approaches used to solve this problem, and different forms of available datasets for sarcasm detection.
\end{abstract}

\def\thefootnote{*}\footnotetext{ Denotes Equal contribution}\def\thefootnote{\arabic{footnote}}

\section{Introduction}

Sarcasm is a communication phenomenon that has received a great deal of attention in the field of linguistics. It is frequently regarded as a complex usage of language in which one communicates the opposite of what they mean. In linguistics, sarcasm has several representations and taxonomies. Wilson \cite{wilson} describes that sarcasm emerges when there is a situational disparity between text and contextual information. Sarcasm is a type of negation in which an explicit negation marker is missing, as per Giora \cite{giora}.

The phrasing of Sarcasm goes against the laws of simple Natural Language Interpretation. Hence, to detect Sarcasm, one needs to understand the syntax of Sarcasm. Sarcasm is primarily built up of a contrast in sentiments, or a disparity between communication and situation. For instance, if someone says “I love being ignored”, the contrast between love and the emotion associated with being ignored explains the presence of sarcasm in the statement \cite{joshi_2}. Given the emotion behind being ignored is subjective, sarcasm becomes hard to detect. Even for humans, Sarcasm has a low average accuracy of 81.6\% \cite{khodak}. Hence, there is no one way of looking at sarcasm - and in turn to solve it.  

Prior to coming up with a solution, a dataset is chosen, taking into account the aspirations of the approach (e.g. multimodal) or the efficacy of the model. For the sake of bench-marking accuracy, using famous databases helps. In some cases, the same code is run on different datasets to understand the circumstances under which the code performs better, and to also ensure that the model is not overfitting to one dataset.

Data is considered for sources - usually Reddit, Twitter, debate data or others, and the form of annotation being conducted. Evaluation metrics usually check for model accuracy, F1 score, precision and recall.

\section{Literature Review}

Sarcasm detection is becoming a popular research topic in Natural Language Processing due to its importance in sentiment analysis and also due to the complexity of detecting sarcasm in text. Joshi et al. \cite{joshi_1} created the first compilation of past work in automatic sarcasm detection. The paper describes the different datasets, approaches, trends and issues in sarcasm detection.
 
\subsection{Linguistic and Context-based Approaches}
Research into sarcasm detection often focuses on the linguistic properties of text and the context in which sarcasm occurs. One prominent approach is based on the linguistic theory of context incongruity. Joshi et al. \cite{joshi_2} developed a system leveraging this theory, demonstrating better performance compared to other systems at the time for both short tweets and longer discussion forum posts. Another context-aware method is proposed by Bamman et al. \cite{bamman}, who introduced a sarcasm detection system incorporating extra-linguistic features, such as the author’s attributes, audience, and the communicative environment, which yielded significant gains in accuracy over systems relying solely on linguistic features.

\subsection{Word Embeddings and Topic Modeling}
Several approaches to sarcasm detection utilize word embeddings to represent textual data. Onan \cite{onan} presented a Deep Learning approach using word-embedding feature sets, specifically employing the LDA2Vec model, which enhances word vector interpretability by linking words to topics. The results indicated that using topic-enriched word embeddings in combination with conventional feature sets produced impressive results in sarcasm detection, particularly on Twitter data.

Agarwal et al. \cite{ameeta} introduced an innovative word-embedding model that integrates affective information into word representations. They found that sentiment affective representations worked best for short texts, like tweets, while more complex representations incorporating fine-grained emotions performed better on longer texts, such as consumer reviews and chat forums.

\subsection{Multi-modal Approaches}
While most sarcasm detection research has focused on text, recent studies have explored multi-modal approaches. Castro et al. \cite{castro} compiled a new dataset comprising audiovisual utterances from popular TV shows annotated for sarcasm, demonstrating that using multi-modal information (e.g., visual and auditory cues) can reduce the error rate in sarcasm detection by 12.9\% in F-score compared to using only individual modalities.

Pan et al. \cite{pan} expanded on this by using multi-modal datasets and Transformer Models. Their approach modeled incongruities between textual inputs and images (inter-modal) as well as between textual inputs and hashtags (intra-modal) in tweets, further advancing sarcasm detection in multi-modal contexts.

\subsection{Graph-based Approaches}
Another emerging area of research involves the use of Graph Networks for sarcasm detection. A study by \cite{graph1} explored a graph convolutional network (GCN) structure to learn inconsistent relationships within and across modalities, using joint and interactive learning to detect sarcasm. Liang et al. \cite{graph2} also utilized a cross-modal GCN, focusing on the identification of inconsistencies between modalities to enhance sarcasm detection.

\section{Sarcasm Datasets}

Sarcasm can be present differently in different modalities. While it can sometimes be present in text without any additional context, it is important to account for the situation, context and common-sense to detect sarcasm. In some scenarios it is hard to understand sarcasm without accounting for the verbal tonation. Lastly, sometimes the sarcasm is multimodal in nature, where solely the text does not suffice to identify the presence of sarcasm. Ideally, since there are multiple types of sarcasm, there need to be different datasets of each type of sarcasm, and different approaches need to be designed to tackle them.

\subsection{Data Sources}

The primary-most sources of datasets for sarcasm were Reddit and Twitter. The commentary-based nature of these applications ensured that opinions were available. And sarcasm is a common form of sharing opinions. Similarly, some data was also taken from Political Debates.

For Multi-modal datasets, one primary source used was Movies or Sitcoms. For instance, the MUStARD dataset used the sitcom F.R.I.E.N.D.S, which is famous for it's apparent sarcasm paired with visual expressions. A complete list of all the sarcasm detection datasets can be found in Table 3.

\subsection{Data Characteristics and Preliminary Analysis}
\subsubsection{MUStARD (2019)}

\begin{figure}
\centering
\includegraphics[width=1.0\textwidth]{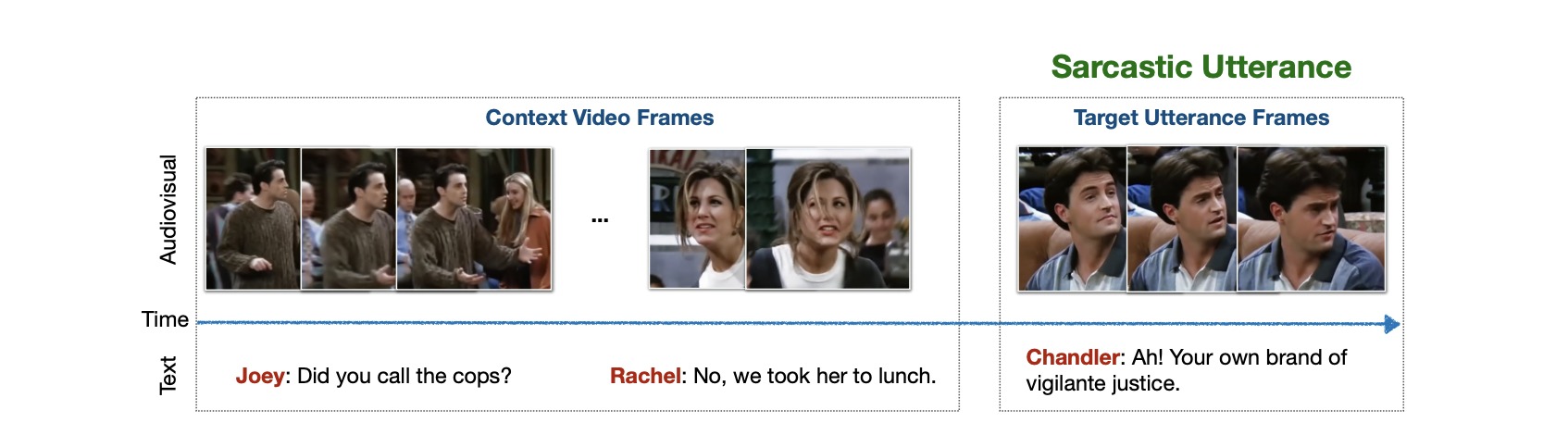}
\caption{\label{fig:Mustard}A sarcastic utterance and its context from the dataset represented by video frames and transcript according to Castro et al. \cite{castro}}
\end{figure}

\begin{figure}
\centering
\includegraphics[width=0.8\textwidth]{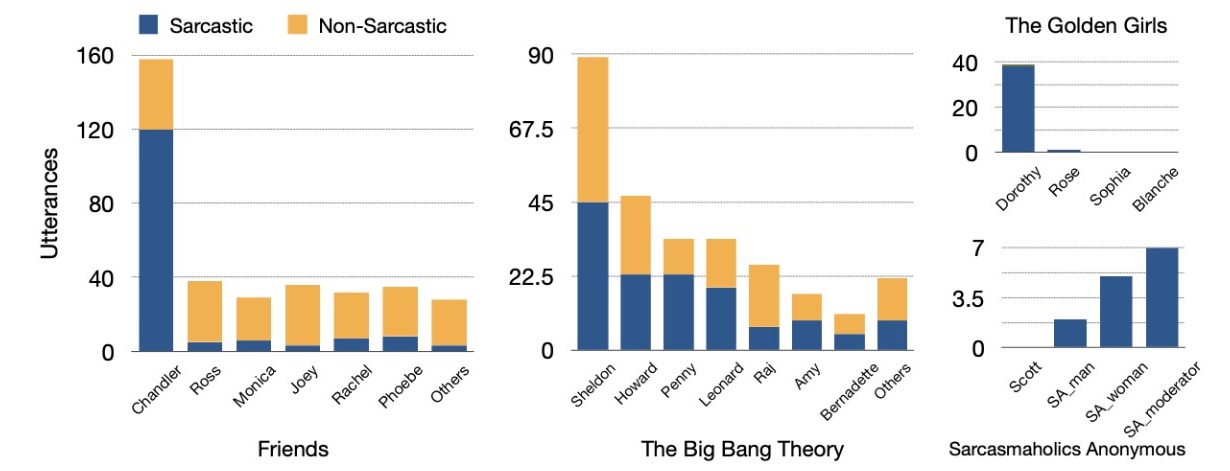}
\caption{\label{fig:graph} Character-label ratio per source according to Castro et al. \cite{castro}}
\end{figure}

MUStARD is a collection of audiovisual utterances annotated with sarcasm labels. To facilitate the research on multi-modal approaches towards sarcasm detection, Castro et al. \cite{castro} proposed this new sarcasm dataset which was compiled from popular TV shows. The videos in the dataset were compiled from four different TV shows: Friends, The Golden Girls, The Big Bang Theory and Sarcasmaholics Anonymous and was then manually annotated. The dataset is constituted of utterances, and each one is accompanied by its historical context in the dialogue, which offers additional information about the situation in which the utterance occurs. Each utterance and its context are comprised of three modalities: video, audio, and transcription (text). Fig. 1 illustrates a sample from the dataset which consists of an utterance and its context represented by video frames and transcription.

Preliminary study of the dataset for sarcasm detection was conducted in the original paper by training an SVM model (SVM models tend to perform better on smaller-sized datasets). While the results indicate a significant reduction in the error rates when using multiple modalities, upon further analysis, we see that the model develops a character bias. With the additional modalities, the model is able to discriminate between a sarcastic person versus a non-sarcastic person (see Fig. 2 for character distribution across the dataset). To test this further, the authors created two test setups — one speaker-dependent and the other speaker-independent. In the speaker-independent setup, there is no overlap of characters between the train and test splits. This setup was only able to achieve a slim reduction in the error rates, but still holds some promise in terms of the potential of multi-modal approaches.
\subsubsection{SARC (2018)}

The Self-Annotated Reddit Corpus, commonly known as SARC \cite{khodak}, is one of the biggest sources of Sarcastic comments from January 2009-April 2017, with ten times more sarcastic comments as compared to other datasets. The total size of the corpus is 533 Million, out of which the percentage of sarcastic comments is 0.25\%, or 1.34 Million. The rate of false positives is 1\% and false negatives rate is 2\% within the dataset.

This dataset employs the self-annotations done through "/s" comments on Reddit, a commonly accepted indication of sarcasm in comment. This ensures that there is a lesser risk of errors during annotations.

To reduce false positives, the dataset ensures that the end of the comment has "/s", and that the user has used "/s" before, suggesting an understanding of the notation.

The SubReddit for Men's rights was the biggest source of sarcastic comments within this corpus.

Conducting an initial test on the dataset gave out the following n-grams that are indicative of sarcasm:
\begin{itemize}
        \item \textbf{Positive Indicator:} obviously, clearly, so fun
        \item \textbf{Negative Indicator:} :), lmao, :(
\end{itemize}

\section{Embeddings}

The word embeddings are important to consider while designing sarcasm detection methods, since the system's understanding of sarcasm is dependent on how it sees the word representations for sarcasm. And there are approaches where a novel method of representing the words to the system can improve its understanding of sarcasm.

\subsection{Topic-enriched (2019)}
Word-embeddings based representation schemes are an important language modelling technique for building deep learning-based schemes for natural language processing tasks. Word-embeddings capture semantic and syntactic relations among words from large sets of documents using unsupervised methods. The commonly used word-embedding models are word2vec, fastText and GloVe.

Word2Vec  has two versions: Continuous Bag-of-words (CBOW) and SkipGram (SG). The CBOW model predicts the central word from a window of words surrounding it, whereas the SG model predicts the context from the central word. The fastText model is an extension of the word2vec model, which generates good representation schemes for rare words while being computationally efficient. The global vectors (GloVe) model aims to integrate word prediction models with word statistics across an entire corpus. Onan \cite{onan} used the LDA2vec word-embedding model based on word2vec for sarcasm detection and compared the performance with the conventional models like word2vec, fastText, and GloVe. LDA2vec allows for the identification of topics in texts and the generation of topic-based word vectors. By trying to link each word to the associated topic, the interpretability of the word vectors has been improved.

The LDA2Vec-based word embedding scheme outperforms other word-embedding-based schemes such as word2vec, fastText, and global vectors in terms of predictive performance. GloVe-based word embedding provided the second best predictive performance in terms of F-measure, while word2vec-based word embedding provided the lowest predictive performance.

\subsection{Affective Representations (2018)}
Agrawal et al. \cite{ameeta} proposed an Affective Word Embedding System(AWES) that uses sentiment and emotion-rich word representations for detecting sarcasm in the text. Here the two spectra of affect that is emotion and sentiment have been exploited. Sentiment consists of binary labels such as positive and negative, and emotion consists of the six categories in Ekman’s model of emotions.

\begin{figure}
\centering
\includegraphics[width=1.0\textwidth]{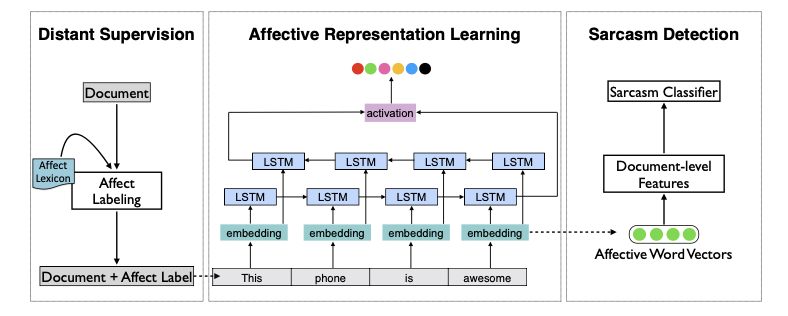}
\caption{\label{fig:affective}An overview of the AWES framework according to Agrawal et al. \cite{ameeta}}
\end{figure}

Words with similar orientations are placed in the same neighborhood in the embedding space. Then distant supervision is used to process the context of individual words in each tweet to automatically label two corpora of product reviews. Since both left and right context of surrounding words in the corpora can contain useful contextual information, BLSTM (Bidirectional LSTM) is used to model to capture from both left to right and right to left. The output of the BLSTM is then flattened and connected to the output layer which is used to predict the target label. Here two models of AWES are being used, one for capturing sentiment information along a binary dimension and the other for encoding a richer spectrum of encodings. Finally the loss for AWES corresponding to sentiment was evaluated using binary cross-entropy where the classification result was passed through softmax to get final labels and, in the case of emotions, loss was evaluated using multi-cross-entropy where the target labels were one-hot-encoded. Fig. 3 illustrates an overview of the AWES framework.

An interesting conclusion was drawn that sentiment-aware representations are most effective for short text sarcasm detection and emotion-aware representations are most effective for detecting sarcasm in longer text.

\section{Approaches}

There is a vast distinction within the approaches for sarcasm detection since sarcasm presents itself uniquely in different cases. This showcases the ingenuity required to detect sarcasm, which will keep evolving even as new approaches for sarcasm detection are designed. A complete list of all the approaches and their details can be found in Table 4.

\subsection{Harnessing context incongruity for sarcasm detection (2015)}
Sarcasm detection research can expand the scope by leveraging well-studied linguistic theories. Joshi et al. \cite{joshi_2} presented a sarcasm detection system that uses a feature set derived from the application of linguistic theory of context incongruity. According to this theory, there are two degrees of incongruity in sarcasm: explicit incongruity and implicit incongruity.

Explicit incongruity is openly expressed through sentiment words of both polarities (as in 'I love being ignored,' which contains both a positive and a negative word). In contrast to opposing polar words, an implicit incongruity is expressed through phrases of implied sentiment. "I adore this paper so much that I made a doggy bag out of it," for example. There is no obvious incongruity here: the only polar word is 'adore'. Nevertheless, the clause 'I made a doggy bag out of it' contains an implied sentiment that contradicts the polar word 'adore.'

In case of explicit incongruity, the features used are the number of times a word is followed by a word of opposite polarity, length of largest sequence of words with same polarity, number of positive words, number of negative words, and lexical polarity of a tweet. In case of implicit incongruity, the feature is a Boolean which indicates the sentiment of implicit phrases. Using these incongruity features in addition to lexical and pragmatic features resulted in an improvement of 8\% in F-score over the models trained with just lexical and pragmatic features.

\subsection{Contextualised Sarcasm Detection (2021)}
To detect sarcasm in tweets, Bamman et al. \cite{bamman} presented a method of creating features out of extra-linguistic information such as the properties of author, audience, and communicative context of the tweet. Extra-linguistic information can be used to generate three types of features in addition to the features of the tweet being predicted. Features generated from the author of that tweet, including historical data by that author; features from the the addressee of the tweet, including historical data for that individual and the historical interaction between the author and the addressee, and features that consider the interaction between the tweet being predicted and the tweet to which it is responding. Five feature combinations were considered to compare the performance of these various feature sets:\\
\begin{enumerate}
        \item Tweet Features
        \item Tweet Features and Response Features
        \item Tweet Features and Audience Features
        \item Tweet Features and Author Features
        \item All the above features
\end{enumerate}
Table 1 shows the different features used in different contexts and Fig.4 illustrates the accuracy achieved over different feature sets. Tweet-only features give an average accuracy of 75.4 \%; adding response features increases this to 77.3 \%; audience features increase this to 79.0\%; and author features increase this to 84.9 \%. Including all features results in the best performance i.e., 85.1 \%, but the majority of these gains are due to the addition of author information.\\
\begin{table}
\begin{tabular}{p{4cm}p{3.5cm}p{3cm}p{3cm}}
 \hline
  Tweet & Author & Audience & Environment\\
 \hline
 Word unigrams and bigrams  & Author historical salient terms & Author/Addressee interactional topics& Pairwise Brown features between the original message and the response\\
 Part of speech features & Author historical topics & Historical communication between author and addressee& Unigram features of the original message\\
 Pronunciation features & Profile information & &  \\
 Tweet whole sentiment &Author historical sentiment  & &  \\
 Tweet word sentiment &   Profile unigrams  & &\\
 Intensifiers &   &   &\\
 \hline
\end{tabular}
\caption{\label{tab:features} Feature sets extracted for each context.}
\end{table}

\begin{figure}
\centering
\includegraphics[width=0.8\textwidth]{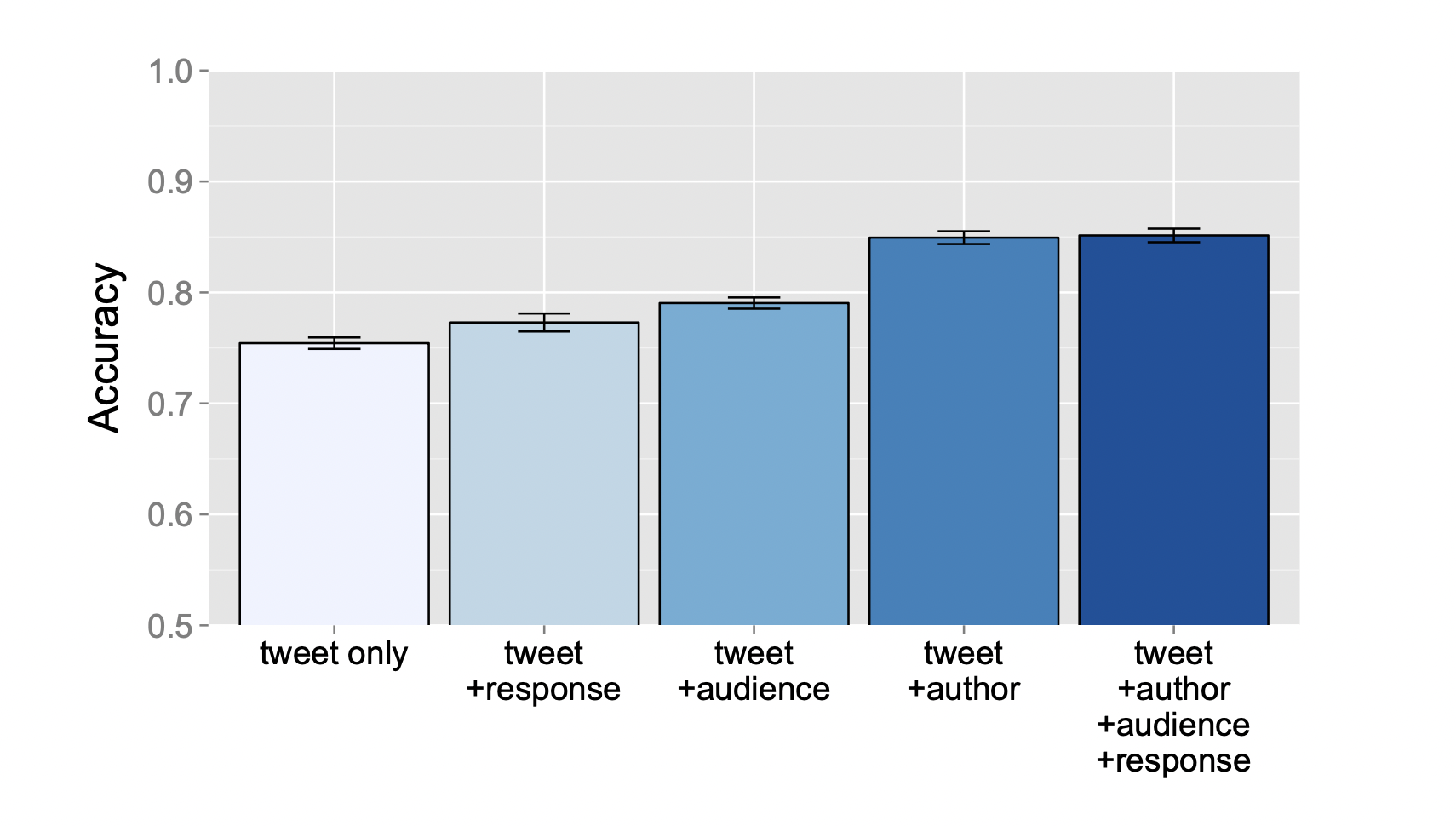}
\caption{\label{fig:context}Accuracy over the five different feature sets according to Bamman et al.\cite{bamman}}
\end{figure}



\subsection{Term weighted Neural Language Models (2021)}

Onan et al. \cite{onan_1} presented a term weighted natural language model and a Deep Neural Network framework for sarcasm detection. Term weighting is a technique to assign appropriate weight to each word or term. To achieve an efficient text representation scheme, an inverse gravity moment based term weighted word-embedding model with trigram features was introduced. The term weighted Neural language model is being integrated into a 3 layered stack BLSTM architecture to identify sarcasm in text documents. This enabled to yield higher predictive performance because richer contextual information was obtained from both past and future. For the evaluation task, the presented framework has been evaluated on three corpus (Twitter messages, ‘‘Sarcasm version 2’’ dataset, ‘‘The News headline dataset for Sarcasm detection’’).
The given scheme was empirically compared with five deep neural architectures (i.e.CNN, RNN, LSTM, GRU, and BLSTM). 

The results indicate that the presented three-layer stacked bidirectional long short-term memory architecture can yield higher predictive performance compared to CNN. The empirical results indicate that the presented word-embedding scheme outperforms the conventional word embedding schemes. The highest value among the compared configurations on the Twitter messages corpus’ has been achieved by fastText trigram-based configuration with inverse gravity moment-based weighting with maximum pooling aggregation, with a classification accuracy of 95.30

\subsection{Reasoning with Sarcasm by Reading In-Between (2018)}

Attention-based models were a revolution in the field of Natural Language, and the Transformer architecture was soon implemented to detect sarcasm better. Prior to this, methods using Gated Recurrent Units (GRU) and Long Short-Term Memory (LSTM) based Models were prevalent. 

Tay et al. \cite{tay} utilizes the research done in theories such as the Situational Disparity Theory (Wilson, 2006) and the Negation Theory (Giora, 1995), and modelled the incongruity and contrast of sarcasm to achieve high accuracy. 

The Multi-dimensional Intra-Attention Recurrent Network model (MIARN), taking inspiration from a self-attention vector, creates an intra-attentive matrix representation of a sentence, where the attention of each word of a sentence is computed against each word of the same sentence. This layer, paired up with a sequential composition LSTM layer, to maintain the understanding of the sentence structure and capture long term dependencies, helps model sarcasm with high accuracy. 

A single-dimensional model with the same architecture was also created (SIARN), and both these models collectively outperformed past models such as GRNN and CNN-LSTM-DNN. The MIARN and SIARN models thus became the new State-of-the-Art model, while maintaining interpretability, as compared to past models. On Twitter Data, MIARN saw about 2\% increase in accuracy and F1 score as compared to prior models.

Potamias et al. \cite{llm} in their paper benchmark the performance of transformer-based LLMs such as BERT, RoBERTa, XLNet in sarcasm and irony detection. These models, thanks to their rich and diverse training, exhibit good transfer learning capabilites and as a result require little data preprocessing and feature engineering when adapting to downstream tasks. The authors also propose a novel architecture combining the pretrained weights of RoBERTa with an RCNN (recurrent convolutional neural net) to improve detection by modelling temporal dependencies of the pretrained embeddings. The experiment results on the Politics subset of the SARC2.0 dataset shows strong performance by transformer models which is further improved by the proposed architecture (see Table. 2).  


[16] introduces SarcPrompt, a prompt tuning method that leverages the task-related knowledge in pretrained language models. Prompt tuning involves formulating the downstream task as a masked language modelling problem, the pretraining objective of many PLMs. In the case of sarcasm, the authors do this by using a prompt template that exploits contradictory intents. Let's take the example of the input phrase 'I love being ignored'. The phrase 'Actually [MASK]' is appended to the input and passed to the model. This phrase is meant to capture the intent of the speaker. If the model predicts the word 'kidding', it indicates contradictory intent for the previous phrase meaning the phrase is sarcastic. The process of selecting "label words" and mapping them to the class labels is called verbalizer engineering. The model is trained to minimize both cross-entropy loss on the class labels as well as contrastive loss on sentence representations to improve distinction between the classes. and The paper elaborates on both the prompt template creation and verbalizer engineering processes in detail.

\begin{table}[h]
\centering
\begin{tabular}{lccccc}
\hline
System & Acc & Pre & Rec & F1 & AUC \\
\hline
ELMo & 0.7 & 0.7 & 0.7 & 0.7 & 0.77 \\
USE & 0.75 & 0.75 & 0.75 & 0.75 & 0.82 \\
NBSVM & 0.65 & 0.65 & 0.65 & 0.65 & 0.68 \\
FastText & 0.63 & 0.65 & 0.61 & 0.63 & 0.64 \\
XLnet & 0.76 & 0.77 & 0.74 & 0.76 & 0.83 \\
BERT-Cased & 0.76 & 0.76 & 0.75 & 0.76 & 0.84 \\
BERT-Uncased & 0.77 & 0.77 & 0.77 & 0.77 & 0.84 \\
RoBERTa & 0.77 & 0.77 & 0.77 & 0.77 & 0.85 \\
CASCADE & 0.74 & - & - & 0.75 & - \\
Ili et al. & 0.79 & - & - & - & - \\
Khodak et al. & 0.77 & - & - & - & - \\
Proposed & 0.79 & 0.78 & 0.78 & 0.78 & 0.85 \\
\hline
\end{tabular}
\caption{Comparison of transformer-based models with RoBERTa-RCNN on SARC2.0 Politics \cite{llm}}
\label{tab:performance}
\end{table}

\begin{figure}
\centering
\includegraphics[width=0.9\textwidth]{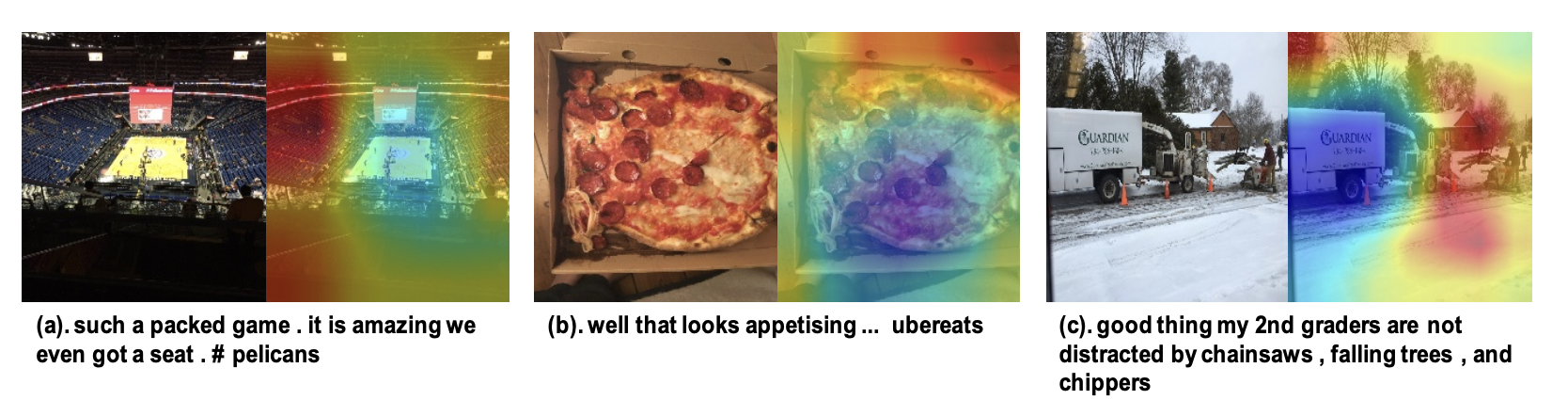}
\caption{\label{fig:multimodal2} How the model looks at multi-modality in sarcasm according to Pan et al. \cite{pan}}
\end{figure}

\subsection{Modeling Intra and Inter-modality Incongruity for Multi-Modal Sarcasm
Detection (2020)}

While MUStARD relies on Sitcoms for it's multi-modal data, Pan et al. \cite{pan} used data from Twitter with images attached. As can be seen in Fig. 5, the inter-modal contrast within the two modalities suggested the presence of sarcasm. Similarly, this approach also targeted intra-modal contrast to detect sarcasm, for instance the incongruity between the text and the hashtag in the Tweet.

The inter-modal modelling is done through a BERT model which takes the input from the Image and text. The image is passed through a ResNet 152 Model before feeding into the Keys and Values of the attention model, while the text input is fed as the query. Then the image and text inputs are matched, attempting to check for incongruities.

On the other hand, the intra-modal modelling happens through a co-attention matrix between the hashtags and the textual input. After this the two dense layers are concatenated, and final predictions are made.

This mechanism surpasses the prior accuracy for approaches applied on multi-modal datasets by 1.25\%. It was also observed that without inter-modal mapping, the model had a accuracy drop of 1.7\%, and without intra-modal mapping, the drop was 0.8\%.

As a further analysis, it was also observed that in case there are beyond three matching layers for image-text during inter-modal modelling, the performance of the model deteriorates.

\begin{figure}
\centering
\includegraphics[width=0.3\textwidth]{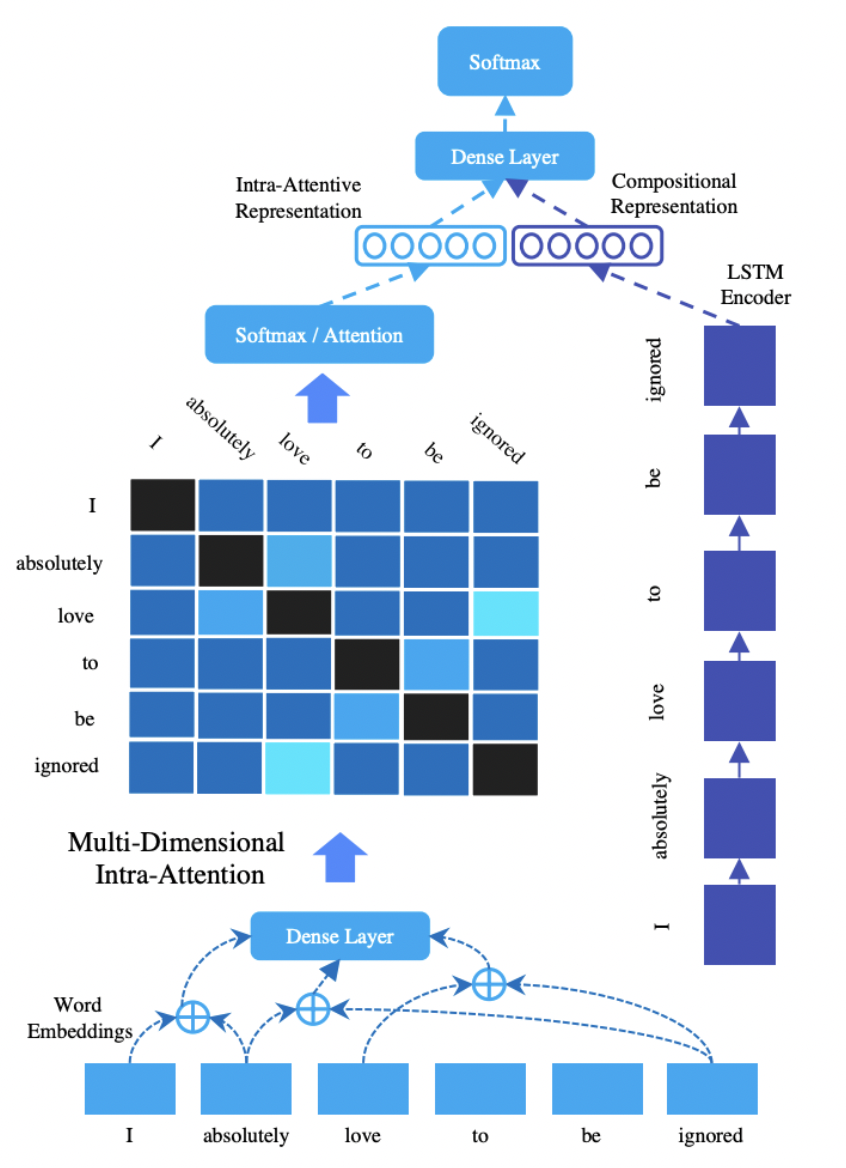}
\includegraphics[width=0.45\textwidth]{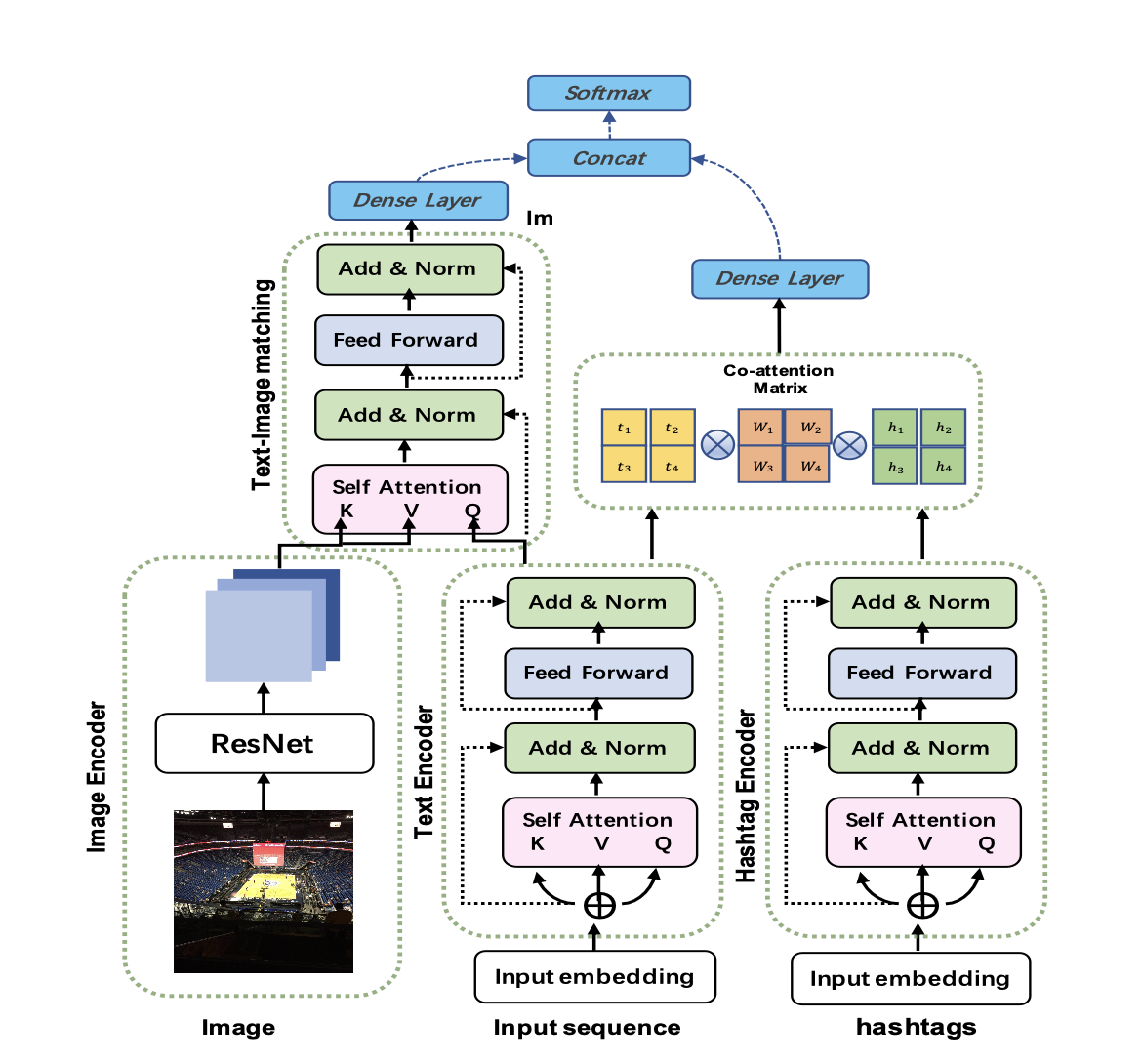}
\caption{\label{fig:rib} Structure of the MIARN model according to Tay et al.\cite{tay} \textit{(left)}. Structure of the multimodal model according to Pan et al.\cite{pan} \textit{(right)}.}
\end{figure}

\subsection{Knowledge fusion network for Multimodal Sarcasm Detection (2023)}

Yue et al. \cite{yue} introduce KnowleNet - a state-of-the-art model for multimodal sarcasm detection by incorporating three novel approaches to sarcasm detection. 

The first module is a Knowledge Fusion network using ConceptNet \cite{speer}, a graph-based semantic network which enables the computer to understand the meaning of words, which is the first model that utilizes prior-knowledge to improve sarcasm-detection accuracy.

Sarcasm has an element of common-sense, especially when we consider multimodality. The image and text would not have any semantic similarity. The second module leverages this ideology using a Multimodal Information Fusion method which checks the semantic similarity between different modes (in this case, image and text), which would be weak in the case of sarcasm.

The third module is a Contrastive-Learning Triplet loss function to improve the representation of the multimodal features to improve the distinction between sarcastic and non-sarcastic samples.

For evaluation, KnowleNet is evaluated on the multimodal sarcasm detection dataset created by Cai et al. \cite{cai}. Cai et al. also processed the dataset to extract image attributes (five descriptive attribute words) from images. Hence, each sample was made up of English Tweets with the corresponding image, and the image attributes.

The model architecture uses four inputs: text, image, image attributes and image caption. The image attributes are from the dataset, while the image captions are generated using the method proposed by Xu et al \cite{xu}. where they use the pre-trained MobileNetV3 to generate captions. 

To encode text and image captions, this paper uses the BERT model, and ResNet was used alongside the Average Pooling operator to encode image data. Next, the ConceptNet model is used to process the text and image attributes to get their vectorized representations and calculate the mutual information between the two. A high value of mutual information would be a signifier of a lack of sarcasm. With the image caption, the spatial distance between the image and text information is used for sample-level similarity detection. 

The Loss function then used is a combination of Binary Crossentropy and Triplet Loss, which aims to minimize the distance between anchor points and positive samples while maintaining the distance between them and negative samples. The KnowleNet approach was compared to other existing models for both unimodal and multimodal inputs. For Text-modality methods: BERT reaches the highest accuracy (83.85\%) and F1-score (80.22\%). Image-modality methods: ConvNeXt and ViT achieve the accuracy of 67.78\% and 67.83\%, which shows text data may contain more effective feature information for Sarcasm Detection. For the Multimodal dataset, KnowleNet performed the best with an Accuracy of 88.87\% and a F1-score of 86.33\%, followed by CMGCN \cite{graph2} with values 87.23\% and 83.45\%.

Lastly, to test the effects of different modules of the model, an ablation study was conducted, where different parts of the model architecture were omitted to see what approach contributed the most to the model, and whether there were any insignificant parts of the model. The results concluded that the mapping and sample-level semantic similarity detection module contributed most significantly to the model, and there was no module that did not positively contribute to the model.

\begin{figure}
\centering
\includegraphics[width=0.6\textwidth]{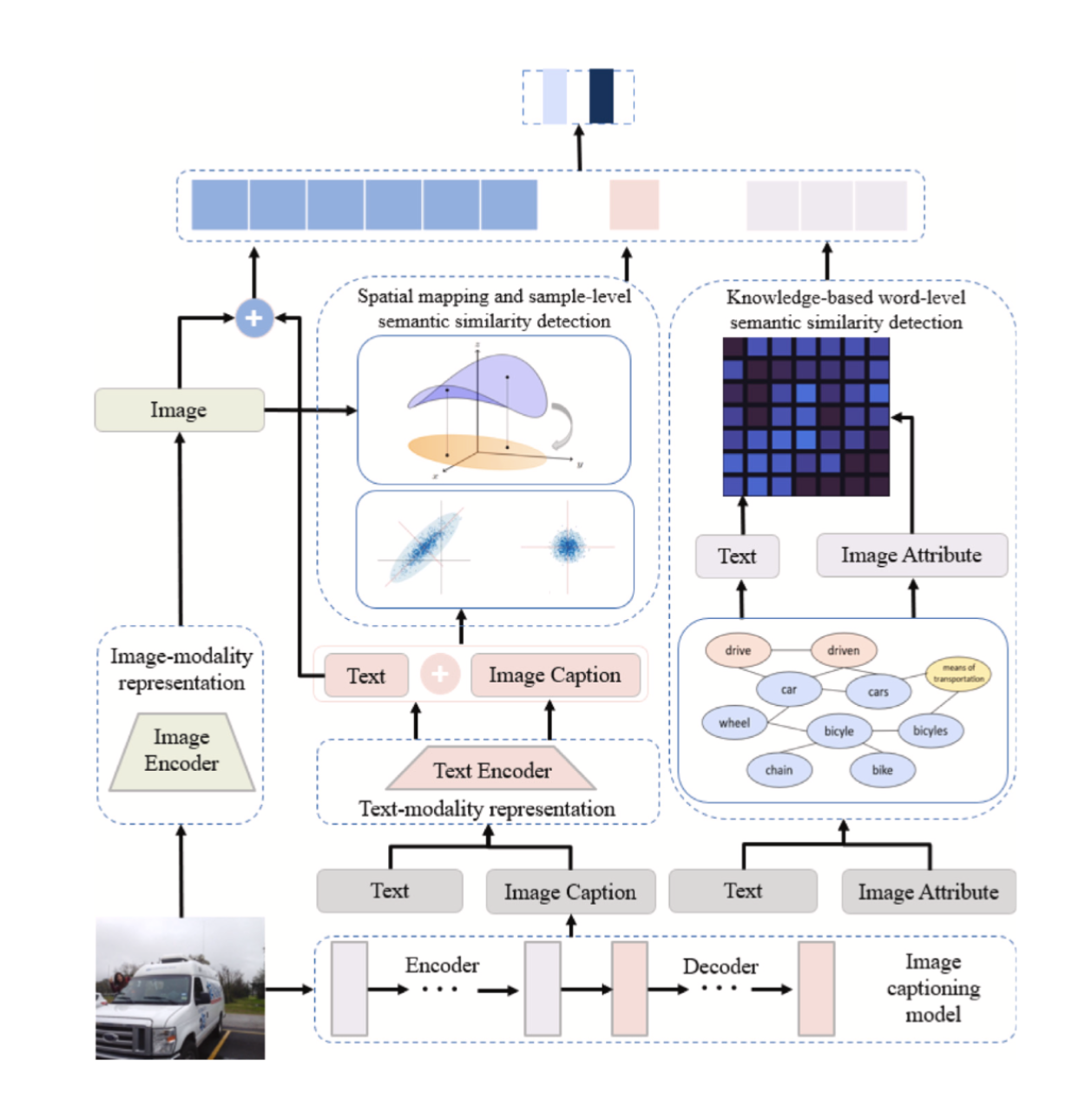}
\caption{\label{fig:knowlenet} The model architecture for KnowleNet in Yue et al. \cite{yue}}
\end{figure}

\section{Future Work}

In the realm of linguistics, the detection of sarcasm presents a multifaceted challenge, continuously evolving with the complexity of its expression. Recent advancements in Generative AI have underscored the performance of large language models (LLMs) for different natural language tasks. A fine-tuned LLM hence holds promise for enhancing sarcasm detection capabilities in text. Moreover, the integration of GPT-Vision could improve upon the state-of-the-art in multimodal sarcasm detection.

In addition, as resources become more readily available for multilingual studies, there is a growing need to extend sarcasm detection efforts to other languages. This entails exploring how sarcasm manifests in different languages, which might also contribute to improving the performance of sarcasm detection in the English language.

Another area of work in sarcasm detection research is the expansion and refinement of existing datasets. One potential approach is generating synthetic samples of Sarcasm to add to the datasets. This approach could improve the robustness and generalization capabilities of the existing sarcasm detection systems.

Lastly, the KnowleNet paper \cite{yue} noticed that metaphors frequently appear in sarcastic expressions, because they allow for a layer of indirectness and irony. Hence, future work might entail checking the presence of metaphors in text to detect sarcasm in a sentence.


\begin{landscape}
\begin{table}[h] 
\begin{tabular}{p{5cm}p{5cm}p{5cm}p{2cm}p{4cm}}
\hline
Name &         Size &         Source &         Modality &         Accuracy Metrics \\ 
\hline
MUStARD: Multi-modal Sarcasm Detection in TV Shows (Castro, 2019) \cite{castro} &         Total number of labeled videos: 6,365 
Number of sarcastic videos: 345 (5.4\% of the total size) &         TV shows (Friends, The Golden Girls, The Big Bang Theory and Sarcasmaholics Anonymous) &         Video, audio, transcription &         F score: 62.8\% (using SVM) \\ 
SARC: Self-Annotated Reddit Corpus (Khodak, 2017) \cite{khodak} &         Total size of the corpus: 533 Million
Number of sarcastic comments: 1.34 Million (0.25\% of the total size) &         Reddit comments &         Text &         F score: 73.2\% (using Bag-of-Words) \\
Multi-Modal Sarcasm Dataset (Cai et al. (2019)) \cite{cai} &         Total size of the corpus: 24,635
Number of sarcastic comments: 10,560 (43\% of the total size) &         COCO image captioning dataset and ResNet pretrained on ImageNet fine-tuned to predict five attributes for each image. &         multimodal (image+text) &         F score: 80.18 \%
Accuracy: 83.44\% \\ 
FLUTE: Figurative Language Understanding through Textual Explanations (Chakrabarty et al. (2022)) \cite{flute} &         Number of sarcastic comments: 9000 &         GPT3, crowd workers, and expert annotators &         Text (sentence pairs with entail/contradict labels and explanations) &         Accuracy: 91.6\% (using T5)
H Score: 85.3\% (using T5) \\ 
WITS: Why Is This Sarcastic (Kumar et al. (2022)) \cite{kumar} &         Number of sarcastic comments: 2240 &         Hindi-English code-mixed sitcom Sarabhai v/s Sarabhai &         multimodal (image+text) &         BERT Score = 77.67\% \\
\hline
\end{tabular}
\caption{\label{tab:features} Datasets for Sarcasm Detection.}
\end{table}
\end{landscape}

\begin{landscape}
\begin{table}[h] 
\begin{tabular}{p{4cm}p{3cm}p{4cm}p{6cm}p{3cm}}
 \hline
Approach &         Model &         Dataset &         Method &         Accuracy Metrics \\ 
\hline
Term weighted Neural Language Models (Onan, 2021) \cite{onan_1} &         Three layer stacked BiLSTM &         sarcasm corpus 1 (self),
sarcasm version 2 (Oraby et al. (2017)) \cite{oraby},
news headline dataset (Misra (2018)) \cite{misra} &         Term Weighted word embedding model with trigrams. Bidirectional LSTM &         Accuracy = 95.30\% \\ 
Reading In-Between (Tay, 2018) \cite{tay} &         multi-dimensional intra-attention, LSTM encoder &         Tweets (Ptáek et al., 2014) \cite{ptaek},(Riloff et al., 2013) \cite{riloff}
Reddit (Khodak et al., 2017) \cite{khodak}
Debates (Lukin and Walker, 2017) \cite{lukin} Khodak &         Attention-based Neural Model &         Accuracy = 86.47\% 
F1 = 86.00\% \\ 
Intra-Inter Modalities (Pan, 2020) \cite{pan} &         BERT &         Cai et al. (2019) \cite{cai} &         Modelling Context Incongruity using BERT &         Accuracy = 84.33\% 
F1 = 86.18\% \\ 
KnowleNet (Yue, 2023) \cite{yue} &         BERT+ResNet &         Cai et al. (2019) \cite{cai} &         Graph-based semantic network semantic. Multimodal Information Fusion. Contrastive-Learning Triplet loss function. &         Accuracy = 92.69\% 
F1 = 91.21\% \\ 
FLUTE (Chakrabarty, 2022) \cite{flute} &         T5 &         Chakrabarty et al. (2022) (self) &         Figurative Language Understanding using Entailments and Contradictions &         Accuracy (T5 fine-tuned on FLUTE) = 91.6\% \\ 
Sarcasm Explanation in Multi-modal Multi-party (Kumar, 2022) \cite{kumar} &         RNN, Transformer, BART, mBART &         Kumar et al. (2022) (self) &         Sarcasm Explanation in Dialogue (SED). Multimodal context-aware attention and global information fusion module. &         BERT Score = 77.67\% \\ 
Multi-Modal Sarcasm Detection via Cross-Modal Graph Convolutional Network (Liang, 2021) \cite{graph2} &         GCN+BERT &         Cai et al. (2019) \cite{cai} &         cross-modal graph convolutional network to make sense of the incongruity relations between modalities &         Accuracy = 87.55\% 
F1 = 84.16\% \\ 
Multi-Modal Sarcasm Detection in Twitter with Hierarchical Fusion Model (Cai, 2019) \cite{cai} &         BiLSTM &         Cai et al. (2019) \cite{cai} (self) &         Hierarchical Fusion Model to combine images, image attributes, and text &         Accuracy = 83.44\%
F1 = 80.18\% \\
 \hline
\end{tabular}
\caption{\label{tab:features} Benchmarks and Approaches for Sarcasm Detection.}
\end{table}
\end{landscape}

\section{Conclusion}

Across this Survey, we started off by understanding what sarcasm entails, which led to a list of unique Sarcasm datasets, each with its unique understanding of the topic (Table 3). We also saw some implementations which have shown success in detection of sarcasm (Table 4). Over the years, the literature in this field has become more prominent, as novel approaches of sarcasm detection are being applied. With the advent of attention based networks, a new wave on sarcasm detection began, with a fresher perspective on how to view this highly creative literary device.


\end{document}